\documentclass[letterpaper, 10 pt, conference]{ieeeconf}  % Comment this line out if you need a4paper

\IEEEoverridecommandlockouts                              % This command is only needed if 
\usepackage{amsmath, amssymb,bbm,bm,mathtools}
\usepackage[ruled,vlined]{algorithm2e}
\usepackage{comment}
\usepackage{url}

\usepackage{graphicx} % for pdf, bitmapped graphics files
\usepackage{epsfig} % for postscript graphics files
\usepackage{amsmath} % assumes amsmath package installed
\usepackage{amssymb}  % assumes amsmath package installed
\usepackage{textcomp}
\usepackage[format=plain,textfont=footnotesize, 
labelfont=footnotesize,
tableposition=top,justification=justified,belowskip=-3pt,aboveskip=8pt]{caption}
\usepackage{xcolor}

\DeclareMathOperator*{\argmax}{argmax} % no space, limits underneath in displays
\DeclareMathOperator*{\diag}{diag} % no space, limits underneath in displays
\renewcommand{\Re}{\mathbb{R}}

\newcommand{\U}{\mathbf{u}}
\newcommand{\V}{\mathbf{v}}

\newcommand{\e}{\bm{\epsilon}}

\newcommand{\T}{\intercal}

\usepackage{amsmath}
\usepackage{accents}
\newlength{\dhatheight}

\newtheorem{property}{Property}[section]

\renewcommand{\baselinestretch}{0.98}

\title{\LARGE \bf
Shield Model Predictive Path Integral: A Computationally Efficient Robust MPC Approach Using Control Barrier Functions
}

\author{Ji Yin$^{1}$, Charles Dawson$^{2}$, Chuchu Fan$^{2}$ and Panagiotis Tsiotras$^{1}$% <-this % stops a space
\thanks{$^{1}$D. Guggenheim School of Aerospace Engineering, Georgia Institute of Technology, GA.
        E-mail: {\tt\small \{jyin81,tsiotras\}@gatech.edu}}%
\thanks{$^{2}$Department of Aeronautics and Astronautics, Massachusetts Institute of Technology, MA. E-mail: {\tt\small \{cbd,chuchu\}@mit.edu}}%
}

\begin{document}

\maketitle
\thispagestyle{empty}
\pagestyle{empty}

%
% ** START EACH NEW SENTENCE IN NEW LINE **
%

%%%%%%%%%%%%%%%%%%%%%%%%%%%%%%%%%%%%%%%%%%%%%%%%%%%%%%%%%%%%%%%%%%%%%%%%%%%%%%%%
\begin{abstract}
 Model Predictive Path Integral (MPPI) control is a type of sampling-based model predictive control that simulates thousands of trajectories and uses these trajectories to synthesize optimal controls on-the-fly. 
 In practice, however, MPPI encounters problems limiting its application. 
 For instance, it has been observed that MPPI tends to make poor decisions if unmodeled dynamics or environmental disturbances exist, preventing its use in safety-critical applications. 
 Moreover, the multi-threaded simulations used by MPPI require significant onboard computational resources, making the algorithm inaccessible to robots without modern GPUs. 
 To alleviate these issues, we propose a novel (Shield-MPPI) algorithm that provides robustness against unpredicted disturbances and achieves real-time 
 planning using a much smaller number of parallel simulations on regular CPUs. 
 The novel Shield-MPPI algorithm is tested on an aggressive autonomous racing platform both in simulation and using experiments. 
  The results show that the proposed controller greatly reduces the number of constraint violations compared to state-of-the-art robust MPPI variants and stochastic MPC methods.

\end{abstract}

\section{Introduction}\label{sec:Introduction}

\begin{figure*}[!h]
    \centering
    \centerline{\includegraphics[scale = 0.40]{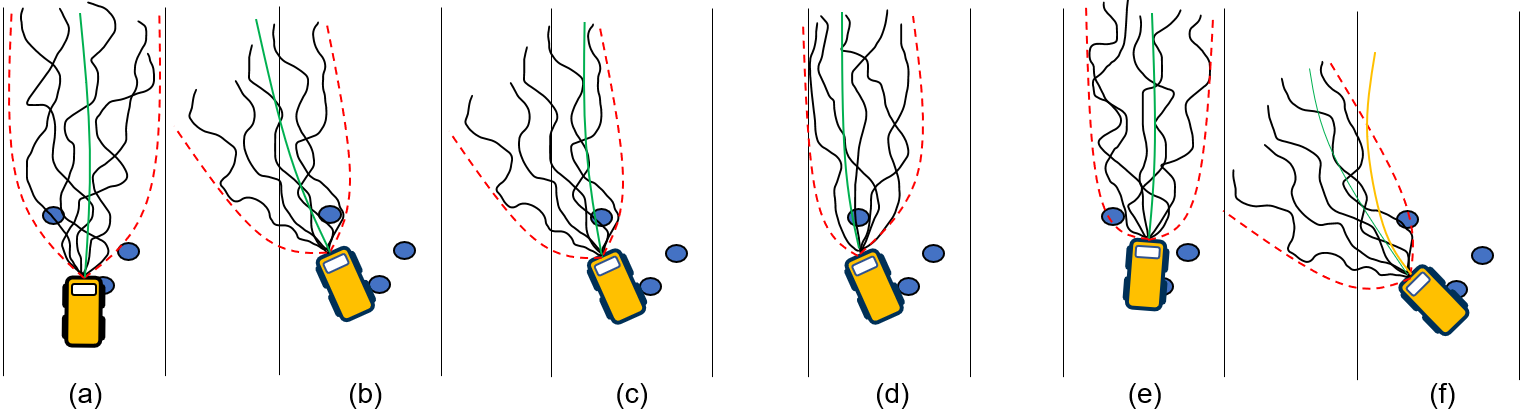}}
    \caption{Comparison of different MPPI variants in the presence of unexpected disturbances. 
    (a)-(b) Environmental disturbances may cause the baseline MPPI to diverge; 
    (c) Some MPPI variants~\cite{UncertaintyPushingMPPI,RAMPPI} penalize trajectories that enter uncertain regions, but provide no guarantees of feasibility; 
    (d) Others variants~\cite{CCMPPI,FlowMPPI,MPPI-CBF} tune the sampling distribution to avoid infeasible states, but these methods can be sub-optimal due to limited exploration or can be biased due to insufficient training data;
    (e) Variants like~\cite{TubeMPPI, RMPPI} pair MPPI with a robust tracking controller, which provides good performance when environmental uncertainty is small but do not formally guarantee safety; 
    (f) The proposed Shield-MPPI guarantees safety even when all trajectory samples deviate from the safe regions, generating feasible solutions shown by the yellow trajectory.}
    \label{fig:MPPIVariants}
\end{figure*}
As robotics technologies develop, autonomous robots are expected to carry out more challenging tasks reliably. To accomplish these tasks in the presence of complex underlying dynamics and unknown environmental conditions, control methods are required to take into account the dynamics along with other user-specified safety constraints. 
Receding horizon control, also known as Model Predictive Control (MPC), is a control method that has been applied to generate optimal controls for constrained robotic systems \cite{CSSMPC}.
Unlike more traditional PID or LQR controllers, MPC considers the future evolution of the system's behavior given the current observation of the states, thus achieving more robust planning \cite{ILQR, snakegait}. We refer the interested reader to \cite{CCMPPI} for a brief review of various categories of MPC algorithms.

Model Predictive Path Integral (MPPI) control is a sampling-based MPC method that relies on forward simulation of randomly sampled trajectories to synthesize an optimal control \cite{KLMPPI}. 
Compared with other MPC approaches, MPPI allows for more general forms of cost functions, including non-convex and even non-smooth costs. 
Typically, MPPI samples a large number of trajectories using a GPU, utilizing the GPU's parallel-computing ability to plan in real time with a sufficiently high control update frequency. 
Despite its attractive properties (e.g., simplicity and support for general nonlinear dynamics and cost functions), MPPI encounters several practical issues when deployed on actual hardware.

First, there exists a gap between the theory of MPPI and its practical implementation. Theoretically, given unlimited computational resources, MPPI will find the globally optimal control sequence, i.e., the algorithm is globally optimal if its planning horizon and trajectory sample budget are infinite. 
In practice, however, the available computational power is always limited. In the past, this problem has been mitigated with the use of GPUs using multi-threaded sampling. 
However, the majority of existing robots still do not have onboard GPUs due to their large size, high cost, and increased power consumption compared to CPUs.

Second, a limited computational budget means that MPPI becomes essentially a local search method. 
As a result, it requires good-quality samples in order to achieve satisfactory performance. Sampling trajectories close to the optimal solutions will significantly improve the performance of the baseline MPPI, just as the quality of initialization affects the performance of any local optimization method. 
A bad set of simulated trajectories with no feasible solutions can cause MPPI to make erroneous control decisions, leading to safety violations. 
In most cases, unexpected dynamical and environmental disturbances cause unsatisfactory behavior, as demonstrated in Fig. \ref{fig:MPPIVariants}(a) and \ref{fig:MPPIVariants}(b). 
In Fig.~\ref{fig:MPPIVariants}(a), the autonomous vehicle has a desirable sampling distribution inside the track, but the vehicle ends up in a state far from the simulated next state due to unexpected disturbances, which may lead to divergence as shown in Fig.~\ref{fig:MPPIVariants}(b).

Third, the baseline MPPI does not consider uncertainty in the environment or the dynamics, and thus neglects potential risks. Specifically, the original MPPI algorithm assumes deterministic dynamics in its trajectory sampling process and imposes a penalty in the cost function as a soft constraint rather than enforcing hard constraints. This use of cost penalties causes two implementation issues. 
First, the cost function has to be carefully tuned and weighted between rewards and penalties, creating the possibility that the algorithm can exploit loopholes in the cost design to make undesirable decisions (so-called ``reward hacking''~\cite{clark_amodei_2019}). 
Secondly, MPPI has no firm guarantees of safety, which can be problematic for many time- and safety-critical applications, such as autonomous driving.

\subsection{Related Work}

Many variants of MPPI have been proposed to address the previous practical limitations. These variants fall into three general categories. 
The first category includes methods designed to address potential planning risks by adding an extra penalty to the sampled trajectories that come close to areas of high uncertainties or risk, pushing the resulting optimal trajectory to high confidence, safer regions, as demonstrated in Fig. \ref{fig:MPPIVariants}(c). For example, \cite{UncertaintyPushingMPPI} uses a data-driven approach to identify uncertainties and avoid potential dangers. 
The authors of \cite{RAMPPI} propose a method to generate risk-averse controls by evaluating the risk in real time and accounting for systematic uncertainties. The major drawback of this category of algorithms is that they may still generate infeasible solutions if none of the sampled trajectories is feasible.

The second category of MPPI variants achieves robust planning by adjusting the distribution of the simulated trajectories to improve sampling efficiency, as described 
in Fig.~\ref{fig:MPPIVariants}(d). 
Reference~\cite{CCMPPI} utilizes covariance steering theory to accomplish flexible trajectory distribution control for MPPI, introducing the final state covariance as a hyper-parameter to adjust the sampling distribution. Other similar methods include~\cite{MPPI-CBF}, which uses a control barrier function to create trust regions for reliable samples, and \cite{FlowMPPI}, which uses a normalizing flow to produce efficient sampling distributions. 
The limitation of these controllers is that their distribution generation method may be biased due to insufficient training data, leading to poor performance. In addition, the constraints on the sampling distribution may limit exploration and lead to sub-optimal plans.

The third category of MPPI extensions addresses systematic uncertainties by closing the gap between MPPI simulations and the actual system~\cite{TubeMPPI, RMPPI} using an additional complimentary controller, such as iLQG, to track the MPPI optimal trajectory, as demonstrated in Fig.~\ref{fig:MPPIVariants}(e). These approaches perform well when the sim-to-real gap is small; 
however, they do not explicitly address risk and they provide no guarantees of safety when the environment changes. 
Such cases are common in autonomous car and drone racing.

\subsection{Contributions}

In this work, we combine control barrier functions with MPPI to develop a safe control approach for general nonlinear systems. Barrier functions are a commonly used verification approach for safety-critical systems that have gained popularity in recent years due to their ability to ensure safety for a wide variety of dynamical systems with safety constraints~\cite{certificate_survey, CBFTA, RBF}.

We integrate the discrete-time control barrier functions (DCBF~\cite{DCBF}) with the MPPI algorithm. 
The resulting Shield-MPPI controller uses a DCBF as a shield to guarantee safety, by filtering the control actions chosen by MPPI to ensure that safety constraints are not violated.
Our approach is inspired from the use of similar safety shields in reinforcement learning~\cite{ShieldRL}, as demonstrated by Fig.~\ref{fig:MPPIVariants}(f). 
The proposed Shield-MPPI possesses two properties that ensure robust planning. 
First, the control actions generated by the Shield-MPPI controller render the specified safe sets forward-invariant, i.e., a Shield-MPPI agent starting inside the safe set will always remain safe. 
Second, if the agent exits the safe set (for example, due to unexpectedly large disturbances), its state will converge back to the safe set, recovering safety as will be discussed in Section~\ref{CBFsection}. 
We will discuss these properties in more detail in Sections~\ref{CBFsection} and~\ref{section: SafetyShields} before providing an experimental characterization of our system in Section~\ref{experiments}. 
In our experiments, the proposed Shield-MPPI controller reduced the chances of a potential car crash to almost zero, while achieving approximately $10-15\%$ speed improvement with less than $0.5\%$ of the trajectory samples used by MPPI.

\section{Model Predictive Path Integral Control}

Consider a general, discrete nonlinear system,
\begin{equation}\label{dynamics}
    x_{k+1} = f(x_k, u_k),
\end{equation}
where $x_k \in \mathcal{D} \subseteq \Re^{n_x}$ is the system state and $u_k \in \Re^{n_u}$ is the control input at time step $k=0,\hdots,K-1$. 
It is assumed that,
given some mean control $v_k\in \Re^{n_u}$ and covariance matrix $\Sigma_\epsilon \in \Re^{n \times n}$,
the actual control follows a Gaussian distribution according to
$u_k \sim \mathcal{N}(v_k, \Sigma_\epsilon)$.
Consequently, the control sequence $\U = (u_0,\hdots, u_{K-1})$ has distribution $\mathbb{Q}$ with density function,
\begin{equation}\label{Qdensity}
    \textbf{q}(\U) = ((2\pi)^{n_u}|\Sigma_\epsilon|)^{-\frac{1}{2}} \prod^{K-1}_{k=0}e^{-\frac{1}{2}(u_k - v_k)^\intercal \Sigma_\epsilon^{-1}(u_k - v_k)}.
\end{equation}
Define the objective function
\begin{equation}\label{objective}
    J(\V) = \mathbb{E}_\mathbb{Q} \left[ \phi(x_K) + \sum^{K-1}_{k=0} \left(q(x_k)  + \frac{\lambda}{2}v_k^\intercal \Sigma^{-1}_\epsilon v_k \right) \right],
\end{equation}
where $q(x_k)$ and $\phi(x_K)$ are the state-dependent step cost and terminal cost, respectively.
As shown in \cite{KLMPPI}, the optimal distribution $\mathbb{Q}^*$ that achieves the minimal value of \eqref{objective} has a density function given by
\begin{equation}\label{optimaldensityfunction}
    \textbf{q}^*(\U) = \frac{1}{\mu}e^{-\frac{1}{\lambda}\left(\phi(x_K)+\sum^{K-1}_{k=0}q(x_k)\right)}\, \textbf{p}(\U),
\end{equation}
where $\textbf{p}(\U)$ is the density function of an (uncontrolled) base distribution $\mathbb{P}$ resulting from a zero-mean control sequence ($\V = 0$), and,
\begin{equation}
    \mu = \int e^{-\frac{1}{\lambda}\left(\phi(x_K)+\sum^{K-1}_{k=0}q(x_k)\right)}\, \textbf{p}(\U)\, \text{d}\U.
\end{equation}

Consequently, the problem of optimizing \eqref{objective} is converted to minimizing the KL divergence between \eqref{optimaldensityfunction} and \eqref{Qdensity}. 
Applying importance sampling, the resulting optimal controls $v_k^+$ can be evaluated using the distribution $\mathbb{Q}$ as,
\begin{equation}\label{optimalcontrol}
    v_k^+ = \mathbb{E}_\mathbb{Q}[u_k w(\U)],
\end{equation}
where,
\begin{equation}\label{eqn:weight}
    w(\U) = \frac{1}{\eta}e^{-\frac{1}{\lambda}S(\U)}.
\end{equation}
and the trajectory cost $S(\U)$ is given by
\begin{equation}\label{eqn:S_cost}
    S(\U) = \phi(x_K)+\sum^{K-1}_{k=0}q(x_k) + \lambda \sum^{K-1}_{k=0} v_k^\T\Sigma_\epsilon^{-1}u_k.
\end{equation}
The denominator $\eta$ in \eqref{eqn:weight} is
\begin{equation}\label{eqn: eta}
    \eta = \int e^{-\frac{1}{\lambda}S(\U)}\, \text{d}\U.
\end{equation}

In practice, \eqref{optimalcontrol} can be calculated using Monte-Carlo sampling as follows. Let $u_k = v_k + \epsilon_k^m$, where $\epsilon_k^m \sim \mathcal{N}(0, \Sigma_\epsilon)$ is the sampled control noise for the $m^\text{th}$ simulated trajectory at the $k^\text{th}$ time step. The control update law \eqref{optimalcontrol} can then be converted to,
\begin{equation}\label{eqn:OptimalControl}
v_k^+ =  \mathbb{E}_\mathbb{Q}[(v_k + \epsilon_k)  w(\U)] \approx  v_k + \sum^{M}_{m=1}\omega_k^m \epsilon_k^m/ \sum^{M}_{m = 1}\omega_k^m,
\end{equation}
where $\omega_k^m$ is the weight for $\epsilon_k^m$ given by \eqref{eqn:weight}, which can be evaluated as,
\begin{equation}\label{eqn:MCweight}
    \omega^m = \text{exp}\left(-\frac{1}{\lambda}\left(S^m - \min_{m=1,\hdots,M}S^m \right)\right),
\end{equation}
where the hyper-parameter $\lambda$ can be used to determine how selective the MPPI algorithm is for the sampled trajectories. For simplicity, in \eqref{eqn:MCweight} we use $S^m$ in place of $S(\U^m)$ to denote the cost of the $m^\text{th}$ simulated trajectory, and the term $\min_{m=1,\hdots,M}S^m$ is introduced to ensure numerical stability without changing the solution. 
It follows from \eqref{eqn:S_cost} that the cost of the $m^\text{th}$ trajectory sample is evaluated as,
\begin{align}\label{eqn:MCTrajectoryCost}
    S^m = \phi(x_K^m) + \sum^{K-1}_{k=0}q(x_{k}^m) + 
     \lambda\, (v_k^m)^\intercal \Sigma_\epsilon^{-1} (v_k^m + \epsilon_k^m). 
 \end{align}

\section{Discrete-time Control Barrier Function}\label{CBFsection}

Let a Lipschitz continuous function $h:\Re^n \rightarrow \Re$, and define a safe set $\mathcal{S} \subseteq \mathcal{D} \subset \Re^n$, such that,
\begin{equation}\label{eqn:safeset}
    \mathcal{S}\coloneqq\{x \in \mathcal{D}| h(x) \geq 0\}.
\end{equation}
Let $\mathcal{U}$ denote the set of feasible controls. 
The function $h$ is a DCBF for system \eqref{dynamics}
if, for all $x\in \mathcal{D}$,
there exists a control $v \in \mathcal{U}$, such that,
\begin{equation}\label{eqn:CBFcondition}
   h(f(x, v)) - h(x) \geq - p (h(x)),
\end{equation}
for a class-$\kappa$ function $p:\Re \rightarrow \Re$.
In this work, we use the specific form of class-$\kappa$ function as follows
\begin{equation}\label{eqn:classKappa}
    p(r) = \beta \, r, \quad \beta \in (0, 1).
\end{equation}

\begin{property}\label{property1}
    Given an initial condition $x_0 \in \mathcal{S}$ and a control sequence $\{v_k\}_{k=0}^\infty$ such that all ($x_k$, $v_k$) pairs satisfy~\eqref{eqn:CBFcondition}, then $x_k \in \mathcal{S}$ for all $k \in \mathbb{Z}_{\geq 0}$.
\end{property}

\begin{proof}
Condition \eqref{eqn:CBFcondition} implies that $h(x_{k}) \geq (\text{Id} - \beta) \circ h(x_{k-1}) $, where $\circ$ denotes function composition and $\mathrm{Id}$
denotes the identity function~\cite{DCBF}. 
Since $h(x_{1}) \geq (\text{Id} - \beta) \circ h(x_0) $, it follows that,
\begin{equation}\label{induction}
    h(x_k) \geq (\text{Id}-\beta)^k \circ h(x_0).
\end{equation}
Since $(\text{Id} - \beta)$ is a class-$\kappa$ function for $\beta \in (0, 1)$, it follows from $h(x_0) \geq 0$ that $h(x_k) \geq 0$. Hence, the set $\mathcal{S}$ is forward invariant.
\end{proof}

\begin{property}\label{property2}
    Let $x_0 \in \mathcal{D}\setminus \mathcal{S}$ and let
     a control sequence $\{v_k\}_{k=0}^\infty$ such that, 
     for all $k \in \mathbb{Z}_{\geq 0}$, the pair
    ($x_k$, $v_k$) satisfies~\eqref{eqn:CBFcondition}.
    Then, the
    state $x_k$ converges to the safe set $\mathcal{S}$ asymptotically.
\end{property}

\begin{proof}
    Note that, as $k \to \infty$, 
    $(\text{Id}-\beta)^k \circ h(x_0) \to 0$. 
    Hence, \eqref{induction} yields $h(x_k) \geq 0$.
\end{proof}

\section{Double-layer Safety Shield using a DCBF}\label{section: SafetyShields}

Integrating safety constraints into an MPPI controller is non-trivial. Since the transition from safe to unsafe states can be abrupt, the controller must consider a sufficiently long planning horizon in order to ensure that the system remains safe far into the future. 
Unfortunately, the need to consider a long planning horizon increases the computation required to evaluate the controller, particularly when the MPPI controller also needs to consider a large number of these long trajectories in order to find near-optimal actions.

\begin{figure*}[!h]
    \centering
    \includegraphics[scale = 0.35]{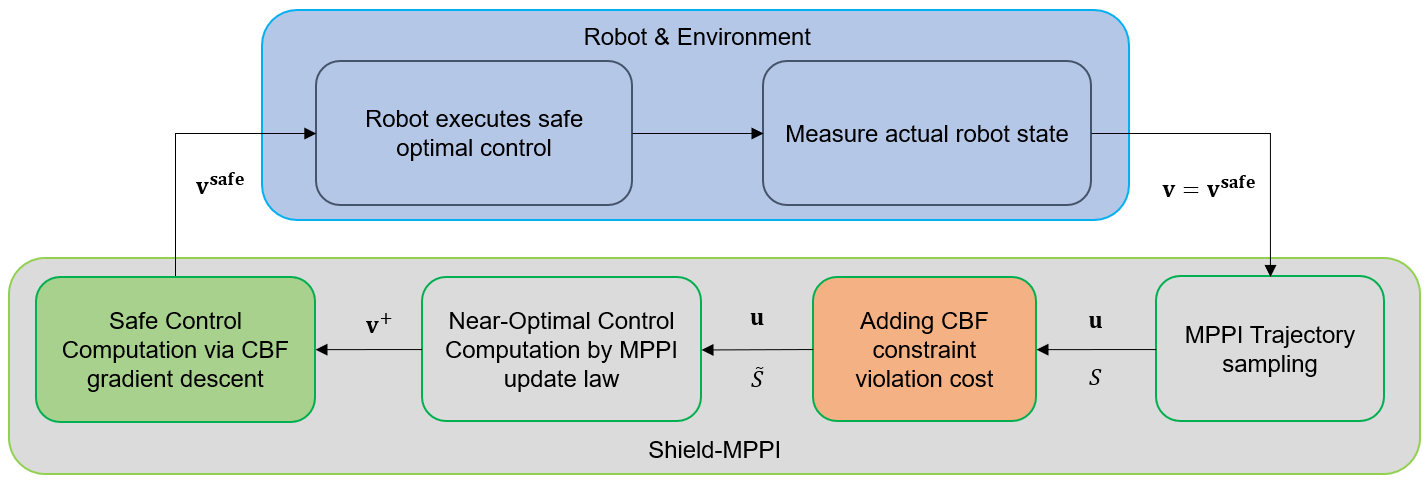}
    \caption{Shield-MPPI control architecture}
    \label{fig:controlframework}
\end{figure*}

To reduce the computational burden required to implement a version of safe MPPI, we make two key modifications to the baseline MPPI controller. 
First, to allow the controller to preserve safety while using a shorter planning horizon, we integrate a control barrier function term into our cost function; this CBF enables the controller to determine whether an action is safe or not, while only considering a handful of steps into the future. 
However, even including a CBF term in the cost may not be enough to ensure safety if the MPPI controller does not consider a large enough sample of trajectories (as this can result in sub-optimal behavior and violation of the CBF's safety guarantee). 
To mitigate this issue and allow the controller to maintain safety even when considering only a small population of trajectories, we combine the CBF-augmented MPPI controller with a local repair step, as shown in Fig.~\ref{fig:controlframework}.

\subsection{Safe Shielding by Modified Trajectory Costs}\label{section:firstshield}

The first component of the proposed control architecture is a standard MPPI sampling process with a state-dependent barrier function term included in the costs of the sampled trajectories. 
To this end, 
let $\alpha = 1 - \beta \in (0, 1)$,
we define a DCBF constraint violation penalty cost, 
\begin{equation}\label{eqn:cbfpenalty:orig}
     C_\text{cbf}(x_k,x_{k-1}) = C\, \max\{-h(x_k) + \alpha h(x_{k-1}), 0\},
\end{equation}
where $C$ is a parameter that determines how much penalty cost should be applied in proportion to the amount of constraint violation. 
In order to augment the CBF constraint into the MPPI cost, we introduce, for each $k=1,\ldots, K$ 
the augmented state $z_k = (z_k^{(1)},z_k^{(2)}) = (x_k,x_{k-1}) \in \Re^{2 n_x}$ and the corresponding augmented state system
\begin{equation}  \label{eq:zeta}
\hspace*{-3mm}
z_{k+1} =
\begin{bmatrix}
z^{(1)}_{k+1}\\[5pt]
z^{(2)}_{k+1}
\end{bmatrix} 
=
\begin{bmatrix}
x_{k+1}\\
x_{k}
\end{bmatrix} =
\begin{bmatrix}
f(z^{(1)}_{k}, u_{k}) \\[2pt] 
z^{(1)}_{k}
\end{bmatrix} = 
\Tilde{f}(z_k, u_k).
\end{equation}
In the new coordinates, equation \eqref{eqn:cbfpenalty:orig} takes the form
\begin{equation}\label{eqn:cbfpenalty}
     C_\text{cbf}(z_k) = C\, \max\{-h(z^{(1)}_k) + \alpha h(z^{(2)}_k), 0\}.
\end{equation}
The new terminal and running costs corresponding to the augmented system~\eqref{eq:zeta}
are then defined as
 $\Tilde{\phi}(z_K) = \phi(z^{(1)}_K) + C_\text{cbf}(z_K)$ and $\Tilde{q}_k(z_k) = q(z^{(1)}_k) + C_\text{cbf}(z_k)$, respectively.
Using the augmented system, the cost of the $m^\text{th}$ simulated trajectory $S^m$ in \eqref{eqn:MCTrajectoryCost} is modified as,
\begin{equation}\label{eqn:ModifiedMCTrajectoryCost}
    \Tilde{S}^m = S^m + \sum_{k=0}^{K} C_\text{cbf}(z^{m}_k),
\end{equation}
where for simplicity, we assume that $z^{(2)}_0 = x_{-1} = x_0$.
If the barrier function constraint \eqref{eqn:CBFcondition} is satisfied it follows that
$-h(x_k) + \alpha h(x_{k-1}) \le 0$ and
the cost term \eqref{eqn:cbfpenalty} becomes zero, and hence the system will remain safe. 
Otherwise, the augmented cost $\tilde{q}_k$ penalizes the simulated trajectories that violate condition \eqref{eqn:CBFcondition}, so that they are weighted less during the synthesis of the MPPI control sequence.

In short, in this step, the MPPI algorithm is applied to system \eqref{eq:zeta} with cost
\begin{align}
    \min_\V J(\V) &= \nonumber \\
    \mathbb{E}&\left[ \Tilde{\phi}(z_K) + \sum^{K-1}_{k=0} \left(\Tilde{q}(z_k)  + \frac{\lambda}{2}v_k^\T \Sigma^{-1}_\epsilon v_k \right) \right],\label{MPPI objective}
\end{align}
to yield a sequence of ``near-optimal'' nominal controls $\V^+ = (v_0^+,v_1^+,\ldots,v_{K-1}^+)$.

\subsection{Control Shielding Using Gradient-based Optimization} 

The MPPI optimization process is not guaranteed to find a solution with zero CBF violation with limited trajectory samples. 
To guard against this case, we add a ``local repair'' step where we seek to locally optimize the output control sequence $\V^+$
 and minimize its violation of the CBF condition, solving the optimization problem,
\begin{equation}\label{eqn:CBFoptimization}
    v^\text{safe}_{0:N} = \argmax_{v^+_{0:N}}  \ \sum_{k=0}^{N} \min\{ h(x_{k+1}) - \alpha h(x_k), 0\},
\end{equation}
subject to \eqref{dynamics}, 
where $x_0$ is the current state and $N$ is the planning horizon for the local repair (typically smaller than the MPPI control horizon $K$). 
If the CBF condition $h(x_{k+1}) - \alpha h(x_k) \geq 0$ is satisfied for $k = 0, 1, \hdots, N$, then the objective of this problem will be $0$, and it will be negative when the CBF condition is not satisfied. We solve this nonlinear problem locally using the Broyden–Fletcher–Goldfarb–Shanno (BFGS) algorithm~\cite{fletcher2013practical}.
The BFGS is a first-order, gradient-based optimizer with a time complexity of $\mathcal{O}(n^2)$, which is significantly faster compared to Newton's method which is of order $\mathcal{O}(n^3)$. 
Due to the real-time constraints on the controller, we do not run this optimization until convergence but instead run it for a fixed number of steps, thus sacrificing any guarantees of local optimality but providing an effective heuristic to ensure safety. 
This approach is illustrated in Algorithm~\ref{algo1}.
\begin{algorithm}[!h]
    \caption{Safety Shield}\label{algo1}
    \SetAlgoLined
    \LinesNumbered
    \SetKwInOut{Input}{Given}
    \Input{
    \noindent Model $f$, repair steps $n_{s}$, MPPI horizon $K$, repair horizon $N < K$, step size $\delta$;}
    \SetKwInOut{Input}{Input}
    \Input{
    Current state $x_0$, control sequence $\V^+$;}
    \SetKwInOut{Output}{Output}
    \Output{
    Safe control $v^\text{safe}_{0:N}$}
    $v^\text{safe}_{0:N} \gets v^+_{0:N}$;\\
    \For{$n_s$ steps}
    {
    $v^\text{safe}_{0:N} \gets v^\text{safe}_{0:N} + \delta \nabla_{v^+_{0:N}}\{ \sum_{k=0}^{N} \min(h(f(x_k, v^+_k)) - \alpha h(x_k), 0) \}$
    }
\end{algorithm}

\section{Shield-MPPI Algorithm}\label{algorithm_section}

The proposed Shield-MPPI is described in Algorithm~\ref{algo2}. 
Line \ref{algo2:GetEstimateLine} computes the estimate of the current system state $x_0$. 
Lines \ref{algo2:TrajectorySamplingBeginLine}~to~\ref{algo2:TrajectorySamplingEndLine} describe the trajectory sampling and cost evaluation process, where Line~\ref{algo2:MPPITrajectoriesRolloutBeginLine} sets the initial conditions, 
Line~\ref{algo2:ControlNoiseSampleLine} samples the $m^\text{th}$ control noise sequence $\e^m$, Line~\ref{algo2:WarmControl} sums the mean control $v_k$ and sampled control noise and Line~\ref{algo2:SystemPropagationLine} uses the resulting input $u_k^m$ to propagate system state. 
Lines~\ref{algo2:TrajectoryCostOriginalLine} and \ref{algo2:TrajectoryCostLine} evaluate the modified trajectory cost  $\Tilde{S}^m$ with the DCBF constraint violation penalty \eqref{eqn:cbfpenalty} following \eqref{eqn:MCTrajectoryCost} and \eqref{eqn:ModifiedMCTrajectoryCost}. 
Line \ref{algo2:CalculateOptimalControlLine} calculates the optimal control $\V^+$ using the update law \eqref{eqn:OptimalControl}. 
To guarantee safety, Line~\ref{algo2:SafetyShieldLine} solves the nonlinear optimization problem \eqref{eqn:CBFoptimization} from Algorithm~\ref{algo1} and obtains the safe control sequence $\V^\text{safe}$. 
Finally, Line~\ref{algo2:ExecuteCommandLine} executes the safe controls and Line~\ref{algo2:InitializationLine} sets $\V^+$ as the mean control sequence for ``warm starting" the next control iteration.
\begin{algorithm}[!h]
    \caption{Shield-MPPI Algorithm}\label{algo2}
    \SetAlgoLined
    \LinesNumbered
    \SetKwInOut{Input}{Given}
    \Input{
    \noindent $\text{Shield-MPPI costs } q(\cdot), \phi(\cdot), \text{parameters} \; \gamma, \Sigma_\epsilon$;}
    \SetKwInOut{Input}{Input}
    \Input{
    \noindent $\text{Initial control sequence } \V$}
    \While{task not complete}{
        $x_{0} \leftarrow \textit{GetStateEstimate}()$;\\ \label{algo2:GetEstimateLine}
        \For{$m\leftarrow 0 \textbf{ to } M-1$ in parallel}{\label{algo2:TrajectorySamplingBeginLine} 
              $x_0^m \leftarrow x_0, \quad z^m_0 \gets [x_0^\T, x_0^\T]^\T, \quad \Tilde{S}^m \leftarrow 0$;\label{algo2:MPPITrajectoriesRolloutBeginLine}\\
              $\text{Sample }\e^m \leftarrow \{\epsilon_0^m,\ldots,\epsilon_{K-1}^m\}$;\label{algo2:ControlNoiseSampleLine}\\
              \For{$k\leftarrow 0 \textbf{ to } K-1$}{
                      $u_{k}^m \leftarrow v_{k} + \epsilon_k^m$;\label{algo2:WarmControl}\\
                      $x_{k+1}^m \gets f(x_k^m , u_k^m)$, \label{algo2:SystemPropagationLine}\\
                      $z_{k+1}^m \gets [(x_{k+1}^m)^\T, (x_k^m)^\T]^\T$ \\
                      $\Tilde{S}^m \leftarrow \Tilde{S}^m + q(x_k^m) + \gamma v_k^\T \Sigma_\epsilon^{-1}u_k^m +~C_\text{cbf}(z_k^m)$;\label{algo2:TrajectoryCostOriginalLine}\\
                      }
                $\Tilde{S}^m \leftarrow \Tilde{S}^m + \phi(x_K^m) + C_\text{cbf}(z^m_K)$;\label{algo2:TrajectoryCostLine}\\ 
        } \label{algo2:TrajectorySamplingEndLine}
        $\V^+ \gets \textit{OptimalControl}(\{\Tilde{S}^m\}_{m=0}^{M-1}, \{\U^m\}_{m=0}^{M-1})$;\\\label{algo2:CalculateOptimalControlLine} 
        $\V^\text{safe} \gets \textit{SafetyShield}(x_0, \V^+)$;\label{algo2:SafetyShieldLine}\\
        $\textit{ExecuteCommand}(v_0^\text{safe})$;\\ \label{algo2:ExecuteCommandLine} 
        $\V \gets \V^+$;\label{algo2:InitializationLine}
    }
\end{algorithm}

\section{Simulation and Experiments}\label{experiments}
In this section, we present simulation and experimental results obtained from running the proposed Shield-MPPI controller on an autonomous racing platform. 
Specifically, we discuss the choice of the DCBF function $h(x)$ along with its corresponding safe set $\mathcal{S}$, and the underlying dynamical system used in these experiments.

\subsection{AutoRally Racing Platform}

We use the AutoRally racing platform~\cite{Autorally}
for simulation as well as experiments. 
The AutoRally is an electric autonomous robot $1/5$ the size of an actual vehicle, which is approximately 1~m in length, 0.4~m in width, and weighs about 22~kg \cite{Autorally}.
We model the dynamics of the AutoRally vehicle using a discrete-time system 
as in \eqref{dynamics},
based on the single-track bicycle model described in \cite{VehicleDynamics},
where system state is $x = [ v_x, v_y,  \dot{\psi}, \omega_F, \omega_R, e_\psi, e_y, s]^\T$, and the state variables represent the longitudinal velocity, lateral velocity, yaw rate, front wheel speed, rear-wheel speed, yaw angle error, lateral deviation, and distance progress made along track centerline, respectively. The control input is $u = [\delta, T]^\T$, where $\delta$ is the steering angle input and $T$ is throttle. 

\subsection{Safe Set}

Assuming that the racing track has constant track width $2 w_{\rm T}$, it is desirable that the vehicle's lateral deviation $e_y$ from the track centerline is bounded by $|e_y| \leq w_{\rm T}$, such that the vehicle avoids collision with the track boundaries. 
To this end, we define the function,
\begin{equation}\label{eqn:DCBFAutoRallyDefinition}
    h(x) = w_{\rm T}^2 - e_y^2,
\end{equation}
that fulfills the DCBF constraint \eqref{eqn:CBFcondition}, that is, $h(x) \geq 0$ if and only if the vehicle is inside the track boundaries.
It follows from \eqref{eqn:safeset} that the safe set $\mathcal{S}$ consists of all states inside the racing track, and Property \ref{property1} indicates that any control policy satisfying \eqref{eqn:CBFcondition} renders $\mathcal{S}$ forward-invariant. 
In addition, Property \ref{property2} guarantees asymptotic convergence to $\mathcal{S}$ in the case when system state is not in $\mathcal{S}$.

\subsection{Controller Cost Design}

In the trajectory cost \eqref{eqn:MCTrajectoryCost}, the state-dependent running cost $q(x_k^m)$ 
can be arbitrary. 
In our simulations and experiments, we used the following state-dependent cost,
\begin{equation}\label{eqn:running_cost}
    q(x_k^m) = (x_k^m - x_{g})^\T Q (x_k^m - x_{g}) + \mathbf{1}(x_k^m),
\end{equation}
where $Q = \diag(q_{v_x}, q_{v_y}, q_{\dot{\psi}}, q_{\omega_F}, q_{\omega_R}, q_{e_\psi}, q_{e_y}, q_s)$ 
are cost weights, $x_g = \text{diag}(v_g, 0, \hdots, 0)$
sets the target velocity, and,
\begin{equation}\label{JC}
    \mathbf{1}(x_k^m)=\left\{
    \begin{array}{ll}
      0, \quad \textrm{if } x_k^m \textrm{ is within the track},\\
      C_\text{obs}, \quad \textrm{otherwise}.
    \end{array}
  \right.
\end{equation}
is the collision penalty cost.

\subsection{Cost Sensitivity Comparison}
\begin{figure*}[!h]
    \centering
    \centerline{\includegraphics[width=0.9\textwidth]{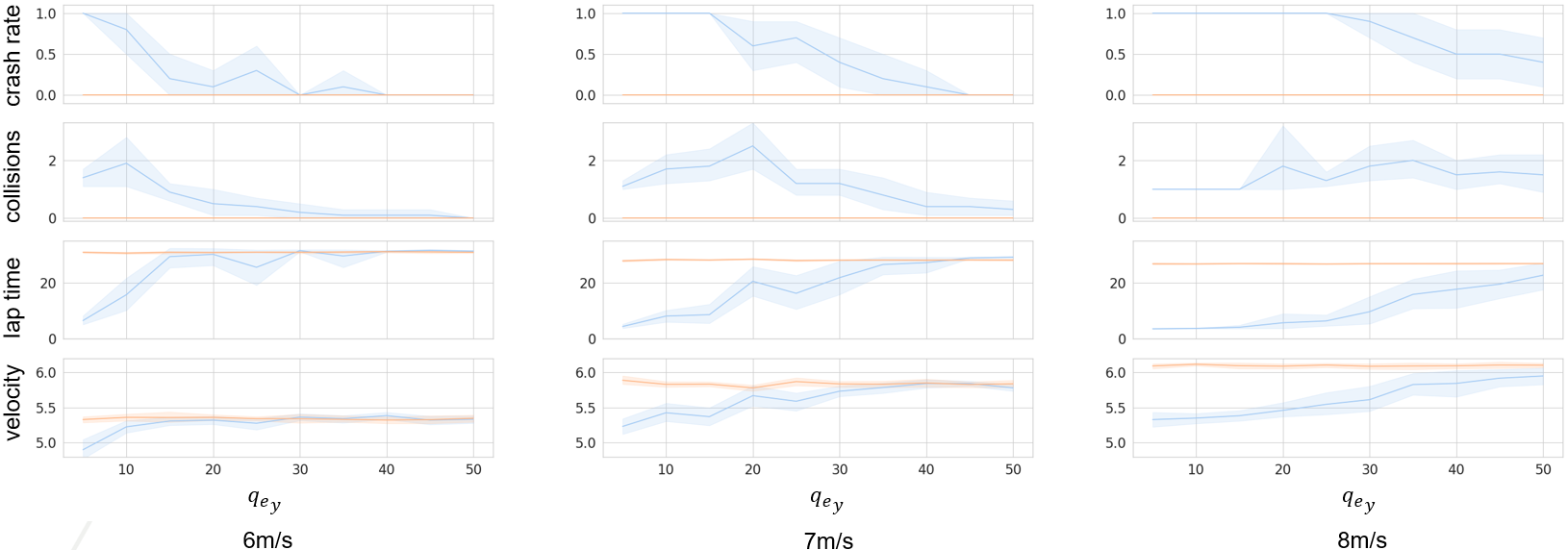}}
    \caption{Cost sensitivity comparison between Shield-MPPI and MPPI.
    Each column is obtained by running the controllers using a different target velocity $v_g$. 
    The blue curves show the performance of the standard MPPI controller, while the orange curves indicate the proposed Shield-MPPI controller. 
    The curves represent average performance with the shaded tubes showing the $95\%$ confidence intervals.}
    \label{fig:LooseCostComparison}
\end{figure*}
A common problem among optimization algorithms is that the cost functions need to be carefully tuned for specific tasks. This is also the case for most MPC controllers, including MPPI. In this section, we investigate the proposed Shield-MPPI's ability to guard against false control decisions made by MPPI by running both controllers using $M= 10^4$ sample trajectories multi-threaded using GPUs.
Normally, the cost weights in \eqref{eqn:running_cost}
% and \eqref{eqn:terminal_cost} 
need to be carefully designed empirically, such that the original MPPI controller achieves satisfying performance.
For the vehicle system \eqref{dynamics}, the cost for a lateral deviation $q_{e_y}$ in \eqref{eqn:running_cost} has a significant impact on the autonomous vehicle's maneuvers. While small $q_{e_y}$ values allow the vehicle to perform aggressive and more time-efficient maneuvers such as cutting corners, a large $q_{e_y}$ makes the system stay close to the track centerline, reducing the chances of a collision against the track boundaries, but at the cost of less efficient trajectories. 
To this end, we tested the original MPPI controller together with the proposed Shield-MPPI controller in simulation, and compared their performance using a wide range of $q_{e_y}$ values. 

\begin{figure}[!h]
    \centering
    \centerline{\includegraphics[scale = 0.23]{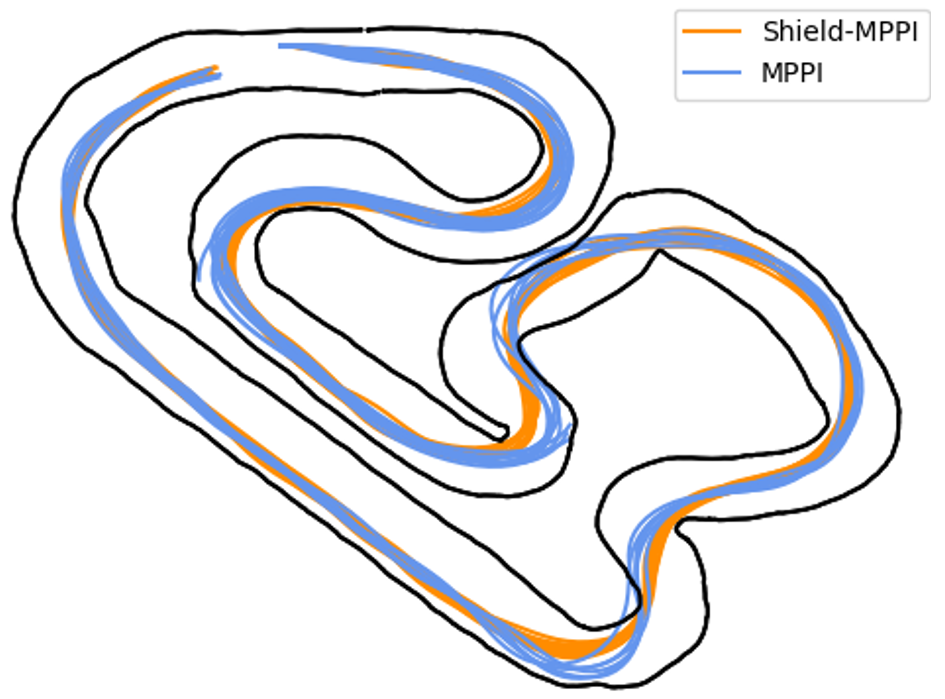}}
    \caption{Shield-MPPI and MPPI trajectory visualization.}
    \label{fig:TrajectoryComparison}
\end{figure}

We define a crash to be the situation where the vehicle deviates far from the track centerline and comes to a complete stop after hitting the track boundaries, and a collision to be the case where the vehicle slightly scrapes the track boundaries but does not halt. 
The first row of plots in Fig.~\ref{fig:LooseCostComparison} shows the crash rates within one lap, and the second row of plots shows the number of collisions. 
The third row illustrates the lap time, which is the time until a crash occurs or the time spent finishing one lap without a crash. The fourth row illustrates the average velocity achieved by the vehicle. 
For a cost interval $q_{e_y} \in [0,50]$, as shown in Fig.~\ref{fig:LooseCostComparison}, the original MPPI's crash rate and the number of collisions increase as the target velocity increases, while the proposed Shield-MPPI controller always maintains zero crash rate and collisions. 
Consequently, the plots in the third row of Fig.~\ref{fig:LooseCostComparison} indicate that MPPI tends to encounter crashes and experience failures earlier than the Shield-MPPI. 
Another observation is that the proposed Shield-MPPI achieves safety with higher velocities than the original MPPI, implying that the proposed approach generates more efficient maneuvers. We visualize trajectories produced by both controllers at $q_{e_y} = 30$ with target velocity $v_g = \text{7m/s}$ in Fig. \ref{fig:TrajectoryComparison}. 
A portion of blue trajectories stops abruptly at the track boundaries, implying crashes caused by MPPI. Some other MPPI trajectories slightly cross the track boundaries and cause minor collisions. 
The trajectories generated by the proposed Shield-MPPI exhibit safe and more efficient driving maneuvers, including cutting corners to avoid losing speed and saving the distance traveled by the vehicle without any collisions.

\setlength{\textfloatsep}{10pt plus 1.0pt minus 2.0pt}

\subsection{Simulations with Limited Computational Resources}

As discussed in Section~\ref{sec:Introduction}, the quality of trajectory samples plays an important role in the optimal control generation for all MPPI-type algorithms. 
Typically, MPPI and its variants rely on the parallel computing abilities offered by modern GPUs to sample as many simulated trajectories as possible to 
find optimal solutions for successful motion generation. 
However, most robots are not equipped with GPUs due to their large size and high cost, and power requirements. 
For this reason,  application of MPPI controllers is restricted to relatively expensive, large-scale robotic systems, while robots designed for affordability and having limited power and size lack the onboard computational resources required to sample a sufficient amount of trajectories in real-time. 
\begin{figure}[!h]
    \centering
    \centerline{\includegraphics[scale = 0.23]{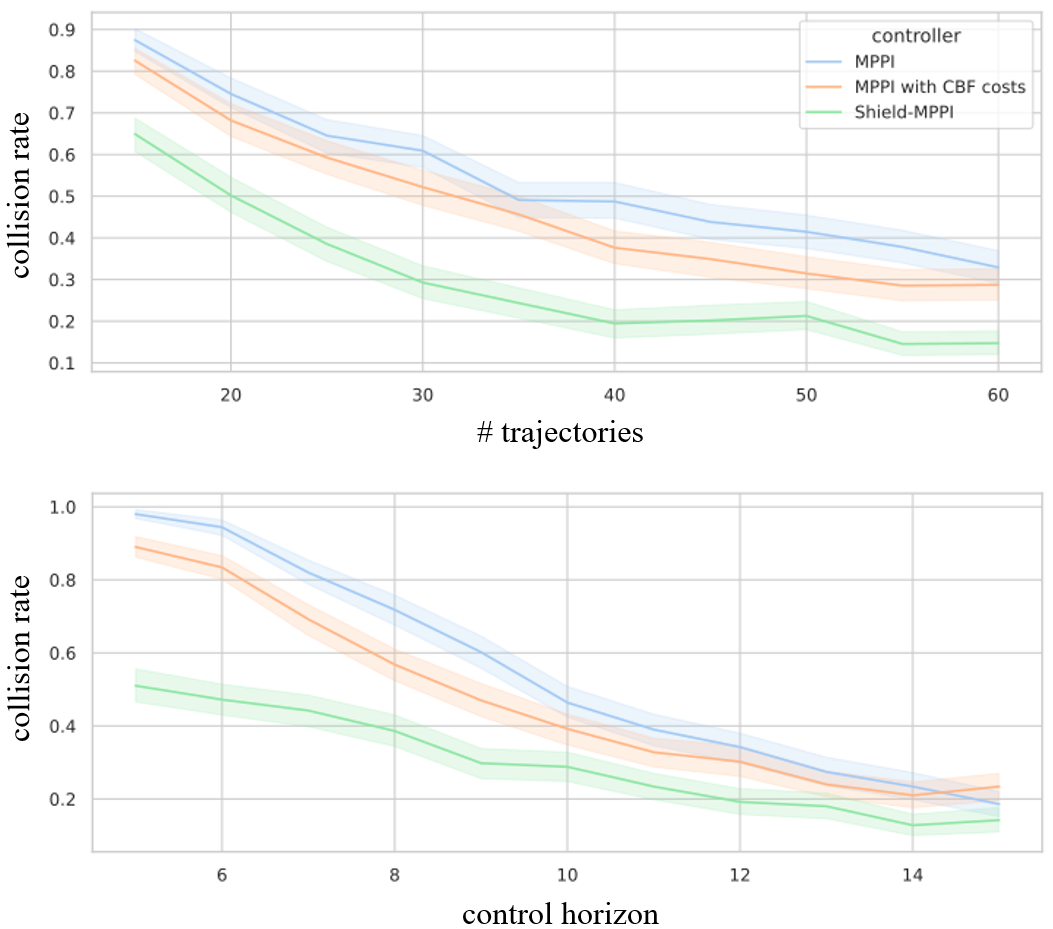}}
    \caption{Comparison of MPPI and Shield-MPPI using CPU implementation.}
    \label{fig:CPUCrashRate}
\end{figure}

To study the proposed algorithm's performance under limited computational resources, we run the original MPPI controller and the proposed Shield-MPPI controller in simulation using as few simulated trajectories as possible with short control horizons on a CPU. 
From the simulation results illustrated in Fig.~\ref{fig:CPUCrashRate}, it is shown that all controllers achieve lower collision rates by increasing the number of trajectory samples and the control horizon. 
The blue curve in the figure indicates that the standard MPPI has the highest collision rate. 
The MPPI using only the DCBF cost modification as described by the orange module in 
Fig.~\ref{fig:controlframework} achieves lower collision rates compared to the standard MPPI, while the proposed two-layer Shield-MPPI, shown in green, results in the minimum number of collisions throughout the entire parameter interval studied.

\begin{figure}[!h]
    \centering
    \centerline{\includegraphics[scale = 0.22]{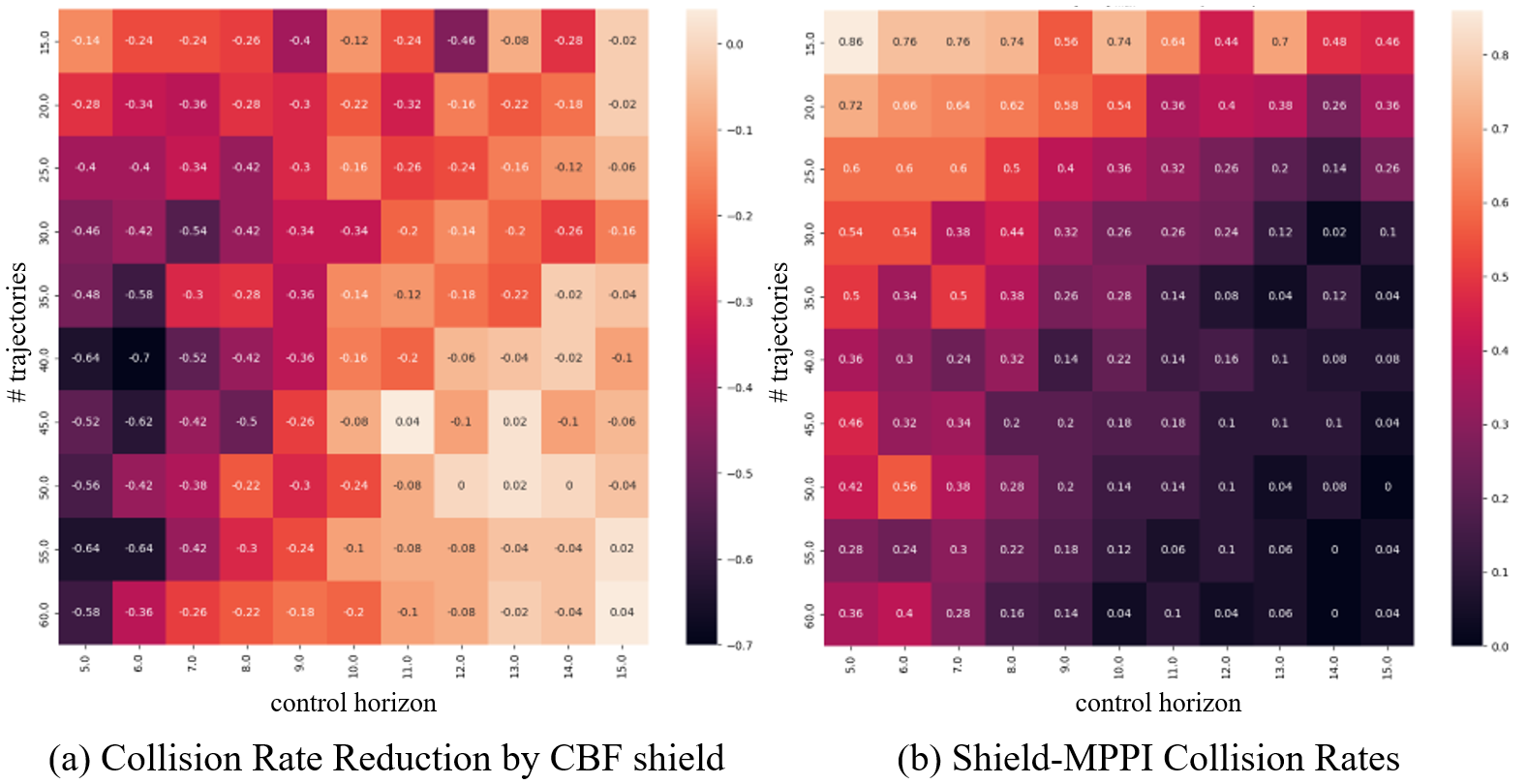}}
    \caption{Collision rate reduction and absolute collision rate of Shield-MPPI controller. 
    Each grid shows the average collision rate reduction or the absolute collision rate over 100 simulations.}
    \label{fig:CrashRateReductionbyShield}
\end{figure}

To further investigate the influence of the second layer safety shield described by Algorithm~\ref{algo1} used in Shield-MPPI, we created a heat map, as shown in 
Fig.~\ref{fig:CrashRateReductionbyShield}(a), to demonstrate the amount of crash rate reduction as a result of Algorithm~\ref{algo1}, using the same data as in Fig.~\ref{fig:CPUCrashRate}. 
The negative numbers in the figure indicate collision rate reduction, where a darker color means more improvement owing to the safety shield. 
It can be observed that the safety shield in Algorithm~\ref{algo1} tends to provide more protection against potential crashes when the control horizon $K$ and the number of trajectory samples $M$ are small, with darker cells appearing in the top-left triangle and lighter ones at the bottom-right corner. 
Fig.~\ref{fig:CrashRateReductionbyShield}(b) shows the absolute collision rates resulting by the proposed Shield-MPPI controller, indicating that the proposed approach achieves zero collisions with merely $50$ samples and about 1.5\,s control horizon.

\subsection{Comparison with other Robust MPC Methods}

To validate the robustness of the proposed Shield-MPPI controller, we compared it with other state-of-the-art controllers that take uncertainties into account during planning. 
In simulations, we model external disturbances by adding some Gaussian noise 
$w_k$ 
to the nominal system \eqref{dynamics}.
It follows that the disturbed system is given by,
\begin{equation}\label{eqn:DisturbedSystem}
    x_{k+1} = f(x_k, u_k) + w_k.
\end{equation}
We ran simulations using the Risk-aware MPPI (RA-MPPI) in \cite{RAMPPI} and the Covariance Steering Stochastic MPC (CS-SMPC) described in \cite{CSSMPC} to compare with our proposed approach.
In addition, we also used a hypothetical Perfect Tracking MPPI (PT-MPPI) that ensures that the actual next state of the agent is the same as the predicted next state from the MPPI, regardless of any disturbances. 
The PT-MPPI assumes perfect trajectory tracking with zero tracking error. 
It is therefore an ideal controller that gives an estimate of the performance upper bound of the tracking-based robust MPPI variants demonstrated in Fig.~\ref{fig:MPPIVariants}(e), including the Tube-MPPI~\cite{TubeMPPI} and Robust-MPPI~\cite{RMPPI}, etc.
To thoroughly test the robustness of the controllers, we use a poor cost design that tends to cause more collisions, and all controllers share the same objective function and control horizon. All MPPI variants sample $10^4$ trajectories at each optimization iteration to ensure a fair comparison. 
Table~\ref{SMPCPerformanceComparison} summarizes the simulation results, which show that the Shield-MPPI controller achieves the lowest crash rate and number of collisions at relatively high velocities.

\begin{table}[!h]
\caption{Performance Comparison with other Stochastic MPC Approaches}
\begin{center}
\begin{tabular}{|c|c|c|c|}
\hline
\textbf{Controller} & \textbf{Crash Rate}& \textbf{Collisions(/lap)} & \textbf{Avg. Speed (m/s)}\\
\hline
$\text{Shield-MPPI}$      & 0.02 & 0.13 & 5.039 \\
\hline
$\text{CS-SMPC}$      & 0.08 & 0.14 & 4.724 \\
\hline
$\text{RA-MPPI}$      & 0.15 & 0.38 & 5.130 \\
\hline
$\text{PT-MPPI}$      & 0.31 & 0.74 & 4.942 \\
\hline
$\text{MPPI}$      & 0.46 & 1.02 & 4.899 \\
\hline
\end{tabular}
\label{SMPCPerformanceComparison}
\end{center}
\end{table}
Another important observation from Table~\ref{SMPCPerformanceComparison} is that while the tracking-based MPPI variants can alleviate the impact of unmodelled disturbances, they are not robust using poorly designed costs due to their lack of risk consideration.

\subsection{AutoRally Experiment}

\begin{figure}[!h]
    \centering
    \centerline{\includegraphics[scale = 0.37]{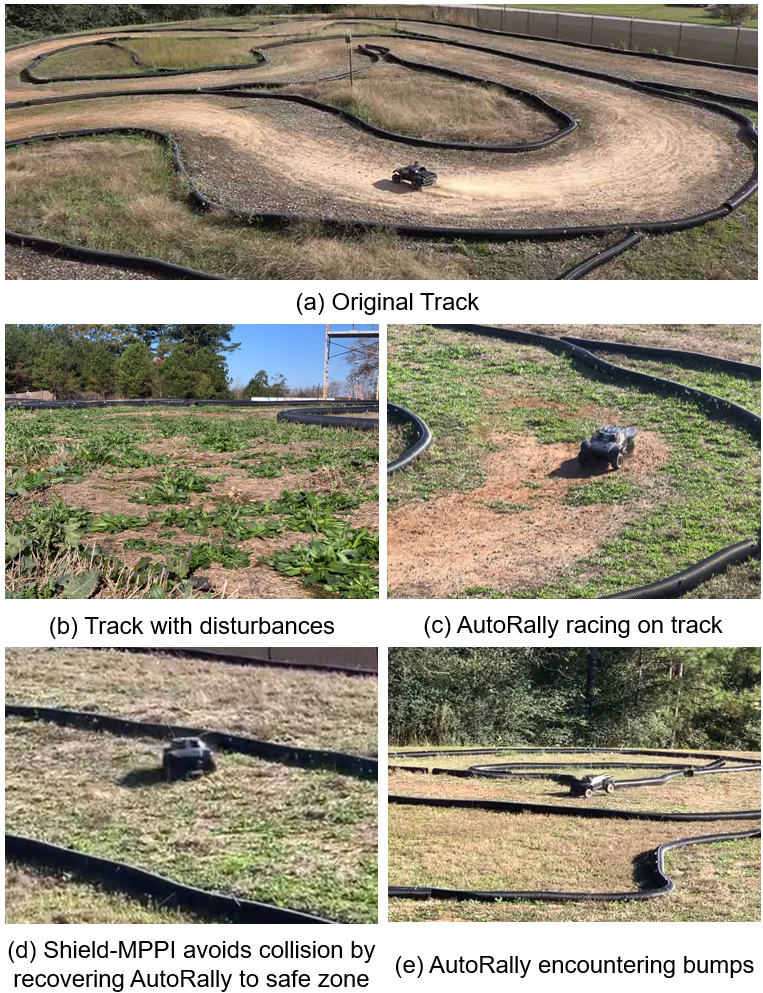}}
    \caption{AutoRally experiment.}
    \label{fig:AutoRally}
\end{figure}

We also investigated the robustness of the proposed Shield-MPPI controller by running it on the real AutoRally vehicle~\cite{Autorally} in the presence of unmodelled external disturbances. 
In our experiments, we tested the controllers on a track subject to disturbances as shown in Fig.~\ref{fig:AutoRally}(b), using a dynamical system \eqref{dynamics} calibrated with the original disturbance-free track as shown in Fig.~\ref{fig:AutoRally}(a). 
Please refer to the video\footnote{\url{https://youtu.be/aKMwEO9wfJ4}} for the experimental demonstration.
The results are summarized in Table~\ref{AutoRallyExperiments}, where the controller MPPI(a) and the Shield-MPPI(a) use GPU to sample $10^4$ simulated trajectories at a frequency of approximately 150 Hz and 55 Hz, respectively, while the MPPI(b) as well as the Shield-MPPI(b) sample 20 trajectories at about 235 Hz and 220 Hz on a CPU. 
The AutoRally vehicle is equipped with the Intel Skylake Quad-core i7 CPU, and an Nvidia GTX 1080ti GPU.
\begin{table}[!h]
\caption{AutoRally Experiment Results}
\begin{center}
\begin{tabular}{|c|c|c|c|c|}
\hline
\textbf{Controller} & \textbf{Samples}& \textbf{Max. Speed (m/s)} & \textbf{Avg. Speed (m/s)} \\
\hline
$\text{MPPI(a)}$      & $10^4$ & 6.31 & 4.30\\
\hline
$\text{Shield-MPPI(a)}$     & $10^4$ & 7.21 & 4.78  \\
\hline
$\text{MPPI(b)}$      & 20 & 4.50 & 2.60  \\
\hline
$\text{Shield-MPPI(b)}$    & 20 & 6.99 & 4.61 \\
\hline
\end{tabular}
\label{AutoRallyExperiments}
\end{center}
\end{table}

From Table~\ref{AutoRallyExperiments}, we see that the proposed Shield-MPPI controller can achieve a $10.78\%$ speed improvement with merely $0.2\%$ the number of trajectory samples compared to the standard MPPI controller, with no collisions observed during the experiments.

\section{Conclusions And Future Work}

In this paper, we proposed the novel Shield-MPPI controller that uses a control barrier function as a shield to prevent unfavorable control performance and guarantee safety. 
In our simulations and experiments, the proposed algorithm significantly reduced the number of safety constraint violations as compared to other state-of-the-art robust MPPI variants and stochastic MPC methods. 
In addition, the Shield-MPPI offers comparable, and even better performance than the baseline MPPI using CPUs instead of expensive GPUs, which has always been a major limitation of applications for MPPI-based algorithms.

In the future, we propose to improve the Shield-MPPI controller using learned certificates as described in~\cite{certificate_survey}, to develop safety shields in more flexible forms, and deploy the resulting algorithms to more complicated control scenarios, such as multi-agent planning~\cite{LearningCertificateMultiAgent}. 
The proposed safety shield in Shield-MPPI can also be integrated with existing MPC methods, such as MPPI variants \cite{RAMPPI, CCMPPI} or robust MPCs \cite{RMPCsurvey}, to further improve their performance and ensure safety.

\section{Acknowledgement}

The authors thank Jacob Knaup for his assistance with the AutoRally platform simulations and experiments. 
This work was funded by NSF under awards CNS-2219755 and CCF-2238030 and by ONR under award N00014-18-1-2828. C. Dawson acknowledges 
support by the NSF Graduate Research Felllowsing Program under grant~1745302.

%%%%%%%%%%%%%%%%%%%%%%%%%%%%%%%%%%%%%%%%%%%%%%%%%%%%%%%%%%%%%%%%%%%%%%%%%%%%%%%%
% \newpage

\bibliographystyle{IEEEtran}
\bibliography{bib/references}

\end{document}